\documentclass{article}

\usepackage[final]{neurips_2026}

\usepackage[utf8]{inputenc} 
\usepackage[T1]{fontenc}    
\usepackage{hyperref}       
\usepackage{url}            
\usepackage{booktabs}       
\usepackage{amsfonts}       
\usepackage{nicefrac}       
\usepackage{microtype}      
\usepackage{xcolor}         
\usepackage{graphicx}
\usepackage{algorithm}
\usepackage{algorithmic}
\usepackage{amsmath}
\usepackage{amssymb}
\usepackage{pgfplots}
\usepackage{CJKutf8}
\usepackage{multirow} 
\usepackage{placeins}
\usepackage[table, dvipsnames]{xcolor}
\usepackage{fancyhdr}
\renewcommand{\headwidth}{\textwidth}

\renewcommand{\headrulewidth}{0.5pt}
\renewcommand{\headrule}{\vspace{2pt}\hbox to\headwidth{\color{black}\leaders\hrule height \headrulewidth\hfill}}
\pgfplotsset{compat=1.18}

\title{X-Cache: Cross-Chunk Block Caching for Few-Step Autoregressive World Models Inference}

\author{\colorbox{white}{\includegraphics[height=0.8em]{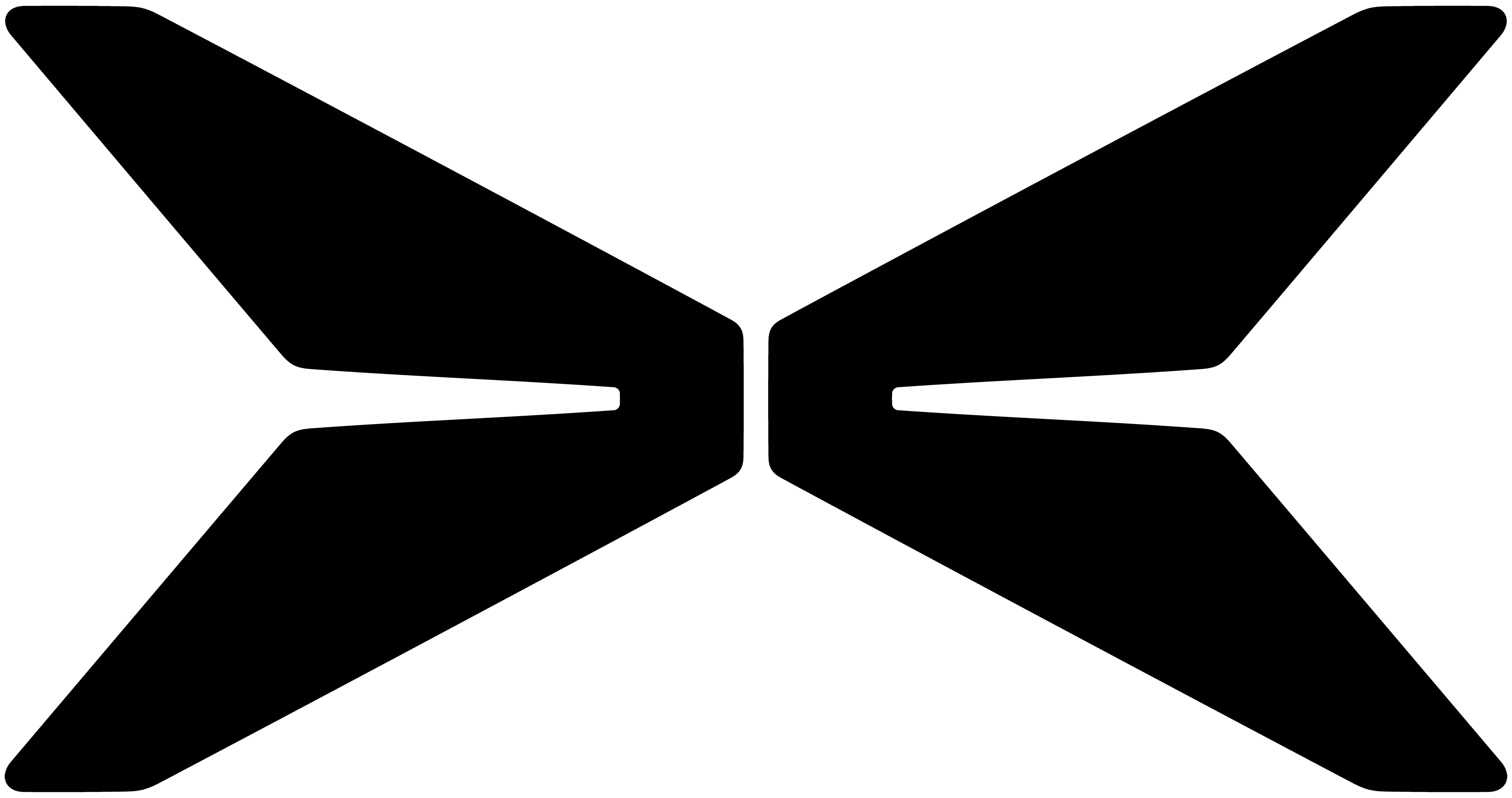}} AI Infra Team, XPeng Inc. \\ 
\textbf{\href{https://x-cache-1.github.io/}{\textcolor{NavyBlue}{https://x-cache-1.github.io/}}}}

\begin{document}

\maketitle
\label{sec:abstract}
\begin{abstract}
    Real-time world simulation is becoming a key infrastructure for scalable evaluation and online reinforcement learning of autonomous driving systems. Recent driving world models built on autoregressive video diffusion achieve high-fidelity, controllable multi-camera generation, but their inference cost remains a bottleneck for interactive deployment. However, existing diffusion caching methods are designed for offline video generation with multiple denoising steps, and do not transfer to this scenario. Few-step distilled models have no inter-step redundancy left for these methods to reuse, and sequence-level parallelization techniques require future conditioning that closed-loop interactive generation does not provide. We present X-Cache, a training-free acceleration method that caches along a different axis: across consecutive generation chunks rather than across denoising steps. X-Cache maintains per-block residual caches that persist across chunks, and applies a dual-metric gating mechanism over a structure- and action-aware block-input fingerprint to independently decide whether each block should recompute or reuse its cached residual. To prevent approximation errors from permanently contaminating the autoregressive KV cache, X-Cache identifies KV update chunks (the forward passes that write clean keys and values into the persistent cache) and unconditionally forces full computation on these chunks, cutting off error propagation. We implement X-Cache on X-world, a production multi-camera action-conditioned driving world model built on multi-block causal DiT with few-step denoising and rolling KV cache. X-Cache achieves \textbf{71\%} block skip rate with \textbf{2.6$\times$} wall-clock speedup while maintaining minimum degradation. 
\end{abstract}
\begin{figure}[H]
    \centering
    \includegraphics[width=0.71\linewidth]{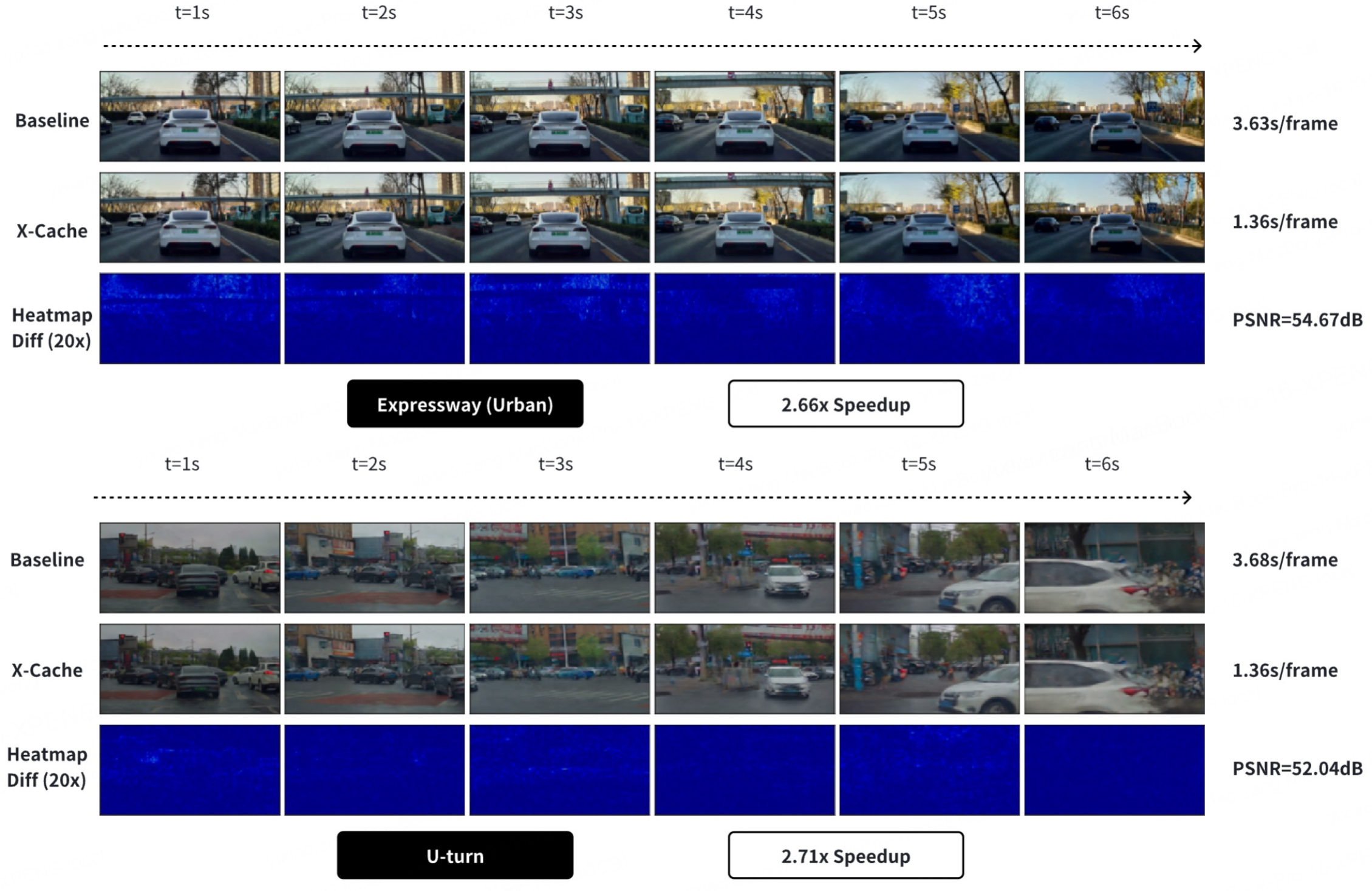}
    \caption{Visual comparison between baseline and X-cache on different simulation scenario}
    \label{fig:overall_comparison}
\end{figure}

\section{Introduction}
\label{sec:intro}

 Generative world models are emerging as a key infrastructure for autonomous driving. By conditioning on ego actions and sensor history, these models generate photorealistic future observations, enabling two capabilities that are difficult to achieve through real-world testing alone: scalable closed-loop evaluation of end-to-end driving policies, and online reinforcement learning where a policy explores counterfactual maneuvers in a controlled, repeatable environment~\cite{x-world,GAIA2,Waymo}. In both settings, the world model must operate as an interactive simulator: generating observations, receiving actions, and responding in real time over long-horizon rollouts.

Autoregressive video diffusion~\cite{causvid,selfforcing,magi1,skyreelsv2} has become a compelling backbone for such simulators. Unlike bidirectional video diffusion that generates an entire clip jointly, autoregressive formulations produce video chunk by chunk with causal attention, conditioning each new chunk solely on previously generated content. This causal, streaming structure naturally supports interactive simulation, where the model must respond to each newly issued action without waiting for a full clip to complete. Combined with a persistent key-value (KV) cache~\cite{ca2vdm,selfforcing,x-world} that bounds memory regardless of sequence length, these models enable stable, unbounded-length generation. To meet throughput demands, they are further distilled to few-step denoising schedules~\cite{dmd,causvid,selfforcing}. This work focuses on accelerating inference for few-step autoregressive video diffusion deployed in interactive, closed-loop world simulation.

A growing line of training-free caching methods~\cite{deepcache,deltadit,teacache,bwcache} accelerates diffusion inference by reusing block outputs in Diffusion Transformers (DiTs)~\cite{dit} across adjacent denoising steps, where structural changes between steps are often small enough~\cite{bwcache,teacache} to permit substantial reuse. Recent extensions to autoregressive video follow the same cross-step design: FlowCache~\cite{flowcache} applies chunk-wise caching policies, while SCOPE~\cite{scope} combines tri-modal scheduling with predictive extrapolation along the denoising trajectory. However, this strategy is poorly suited to the few-step regime required for real-time interactive world simulation. With only four denoising steps, the cross-step similarity that these methods rely on is greatly reduced. Each step contributes substantial and non-redundant structural updates, leaving little room for safe reuse. FlowCache and SCOPE are subject to the same limitation: both are built on the assumption of cross-step redundancy and have only been evaluated under many-step schedules.

Beyond the few-step barrier, accelerating interactive world simulation exposes two structural complications. First, a major subset of training-free caching methods~\cite{teacache,worldcache,scope,flowcache} relies on extrapolating cached feature trajectories under a local smoothness assumption. In interactive simulation, the per-chunk action stream emitted by an external policy is by design non-smooth at chunk boundaries (discrete braking, steering, or lane-change commands enter every DiT block via adaLN-Zero), so the smoothness assumption fails precisely at the chunk boundaries where the world model is supposed to respond to a new control input. Second, specific to interactive simulation, the generation loop imposes a strict causal dependency: the world model must wait for an external policy to observe the fully generated current chunk and output a corresponding action before the next generation step can begin. This sequential dependency rules out sequence-level parallelization techniques (e.g., Block Cascading~\cite{blockcascading}) that rely on knowing future conditionals upfront to concurrently decode multiple chunks, so all acceleration must stay within the current generation boundary.

Despite these limitations, we observe that few-step autoregressive world models exhibit strong redundancy along a different axis: \textbf{across consecutive generation steps} rather than across denoising steps. In autonomous driving the scene evolves slowly relative to the generation rate, so the physical world changes continuously but smoothly between adjacent chunks. DiT block inputs at matching (denoising step, block) positions are then largely similar between consecutive generations. This cross-chunk redundancy comes from physical scene continuity rather than from proximity along the denoising trajectory, so it survives the few-step distillation that erases cross-step redundancy. Based on this observation, we propose \textbf{X-Cache}, a training-free block-level caching method that exploits cross-chunk redundancy for few-step autoregressive world model inference. Our contributions are:

\begin{enumerate}

\item \textbf{Cross-chunk block caching:} We cache DiT block residuals across consecutive generation steps at matching denoising step indices, exploiting temporal redundancy that persists under few-step distillation. To our knowledge, this is the first caching method that operates along the cross-chunk axis.

\item \textbf{Structure-aware, action-aware fingerprint:} To make cross-chunk similarity tractable at every (denoising step, block) position, we introduce a compact fingerprint that subsamples block inputs on the latent's 3D $(F, H, W)$ grid rather than along the flattened 1D token axis, giving geometrically balanced coverage across frames and spatial locations. We further attach a global mean channel to catch bulk drift that the sparse spatial sample may miss, and an action-condition channel that lifts the per-chunk action vector (consumed inside each block via adaLN-Zero) back into fingerprint space. This restores direct sensitivity to per-step control maneuvers that flat input-similarity metrics would otherwise miss, while dynamic-object and lane conditioning, which have unstable padding across chunks, are handled by the step-0 protection and anchor-block mechanisms introduced next.

\item \textbf{KV update frame protection:} We identify the KV cache writing pass as a critical error amplification point unique to autoregressive inference and unconditionally force full computation on these frames, cutting off permanent error propagation.

\item \textbf{Adaptive per-position thresholding:} To avoid hand-tuning a single global similarity threshold across all blocks and denoising steps, we introduce a per-(denoising step, block) adaptive cosine threshold that auto-regulates from the block's own cross-chunk similarity history via an exponential moving average, clamped from below by a global quality floor. Blocks whose cross-chunk similarity is consistently high accumulate aggressive thresholds and contribute most of the skips, while blocks with volatile similarity remain conservative, yielding a stable per-chunk compute profile without any explicit budget controller.

\item \textbf{Validation on a production driving world model:} We validate X-Cache on a multi-camera, action-conditioned driving world model with few-step causal denoising and rolling KV cache. X-Cache achieves 71\% block skip rate with 2.6$\times$ DiT wall-clock speedup at negligible quality degradation in closed-loop simulation.

\end{enumerate}

We note that direct comparison with existing caching methods is not applicable: all prior methods operate along the cross-step axis under many-step, one-shot generation settings, and cannot be directly applied to the interactive, few-step autoregressive world model that X-Cache targets. X-Cache exploits a different redundancy axis entirely, making it complementary to rather than competitive with step-wise approaches. Our experiments therefore focus on ablation studies and absolute performance rather than head-to-head comparisons.
\section{Method}
\label{sec:method}

\begin{figure}[h]
    \centering
    \includegraphics[width=0.9\linewidth]{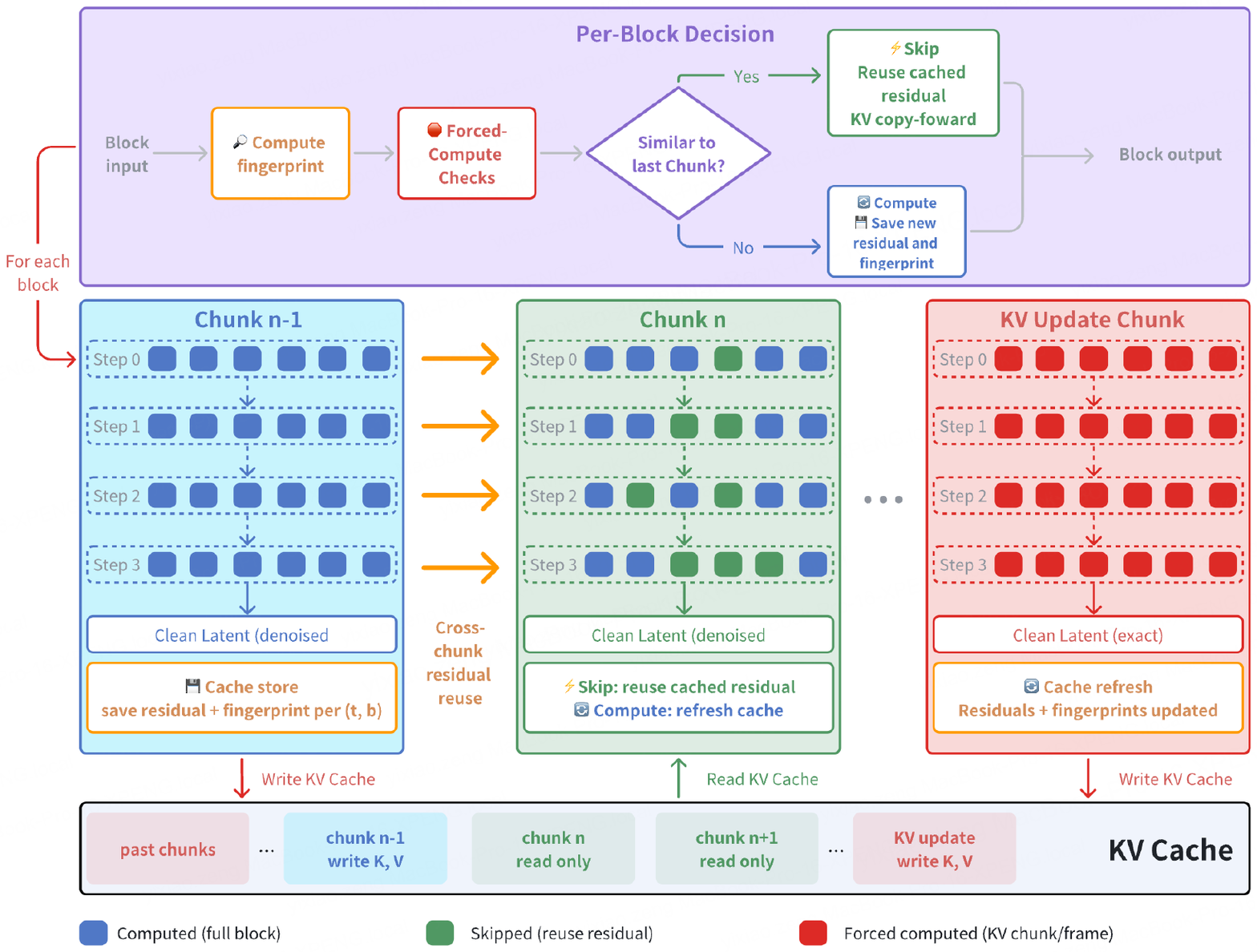}
    \caption{Overall architecture of X-Cache}
    \label{fig:x-cache}
\end{figure}
We first establish notation for the AR video diffusion inference loop, then describe each component of the caching mechanism.

\subsection{Preliminaries: AR Video Diffusion Inference}
\label{sec:method:prelim}

Let $n$ index the generation step (i.e., the $n$-th chunk of video), $t \in \{0, \ldots, S{-}1\}$ the denoising step within a chunk, and $b \in \{0, \ldots, B{-}1\}$ the DiT block index. At generation step $n$, chunk latents are initialized from Gaussian noise and iteratively refined through $S$ denoising steps, where each step passes through all $B$ DiT blocks:
\begin{equation}
\label{eq:dit_forward}
\mathbf{x}^{(n)}_{t,b} = \mathbf{x}^{(n)}_{t,b-1} + f_b\!\left(\mathbf{x}^{(n)}_{t,b-1};\, \mathbf{c}^{(n)}_t\right),
\end{equation}
where $\mathbf{x}^{(n)}_{t,0}$ is the latent at the start of block 0 at denoising step $t$, $f_b(\cdot;\, \mathbf{c})$ denotes the $b$-th DiT block parameterized by conditioning $\mathbf{c}$ (timestep embedding, action embedding, text embedding, etc.), and the operation is a standard residual connection. We define the block residual as:
\begin{equation}
\label{eq:residual}
\mathbf{r}^{(n)}_{t,b} \;=\; f_b\!\left(\mathbf{x}^{(n)}_{t,b-1};\, \mathbf{c}^{(n)}_t\right) \;=\; \mathbf{x}^{(n)}_{t,b} - \mathbf{x}^{(n)}_{t,b-1}.
\end{equation}

After all $S$ denoising steps complete for chunk $n$, the model performs a KV update pass: it runs one additional forward pass through all blocks using the fully denoised (clean) latent, computing key-value projections that are appended to the persistent KV cache for conditioning future chunks via cross-attention. In rolling KV cache implementations, oldest entries are evicted via FIFO when the cache reaches capacity.

\subsection{Cross-Chunk Residual Caching}
\label{sec:method:caching}

\subsubsection{Key observation} 
In AR video generation of physically continuous environments (especially in autonomous driving), consecutive chunks depict scenes that change smoothly relative to the generation rate. The block input $\mathbf{x}^{(n)}_{t,b-1}$ at position $(t, b)$ is therefore highly similar to $\mathbf{x}^{(n-1)}_{t,b-1}$ at the same position in the previous generation step. This cross-chunk redundancy is independent of the number of denoising steps $S$ and persists even under aggressive few-step distillation (e.g., $S=4$), unlike cross-step redundancy, which diminishes as $S$ decreases.

\subsubsection{Caching mechanism} 
When block $b$ is fully computed at generation step $n$ and denoising step $t$, we cache its residual:
\begin{equation}
\label{eq:cache_write}
\hat{\mathbf{r}}_{t,b} \;\leftarrow\; \mathbf{r}^{(n)}_{t,b} = \mathbf{x}^{(n)}_{t,b} - \mathbf{x}^{(n)}_{t,b-1}.
\end{equation}
At the next generation step $n{+}1$, if the gating mechanism (Section~\ref{sec:method:gating}) determines that block $b$ at denoising step $t$ can be skipped, we approximate its output by additive reuse:
\begin{equation}
\label{eq:cache_read}
\tilde{\mathbf{x}}^{(n+1)}_{t,b} = \mathbf{x}^{(n+1)}_{t,b-1} + \hat{\mathbf{r}}_{t,b}.
\end{equation}
The cached residual is indexed by the pair $(t, b)$, ensuring reuse occurs between matching positions in the denoising trajectory.

\subsubsection{Initialization} 
No cached residuals exist for the first generation step ($n=0$). During this \emph{warmup phase} (configurable to $W \geq 1$ steps), all blocks compute fully, populating the cache for every $(t, b)$ pair.

\subsection{Dual-Metric Gating}
\label{sec:method:gating}

Beyond the warmup phase, X-Cache evaluates whether each block can be safely skipped using a dual-metric similarity test between the current input fingerprint and the cached fingerprint from the same $(t, b)$ position at the previous generation.

\subsubsection{Fingerprint} 
Each block input $\mathbf{x}^{(n)}_{t,b-1}$ has shape $B{\times}V{\times}L{\times}C$, where $V$ is the number of cameras in a view group and the token axis $L$ is a flattened $(F_g{\times}H_g{\times}W_g)$ spatio-temporal grid. Comparing the full tensor at every $(t, b)$ is too expensive, and uniform 1D subsampling along $L$ would cover frames and spatial locations unevenly. We instead subsample on the 3D grid: given a target budget of $K$ tokens, we allocate the three axes in proportion to the grid's aspect ratio ($k_F : k_H : k_W \approx F_g : H_g : W_g$ with $k_F k_H k_W \approx K$), take uniformly spaced indices along each axis, and index $\mathbf{x}$ with their Cartesian product, yielding a compact fingerprint $\phi(\mathbf{x})$ of shape $B{\times}V{\times}(k_F k_H k_W){\times}C$ (we use $K{=}32$). For blocks with $L{\le}K$ we keep the full $\mathbf{x}$; if the grid shape is unavailable at runtime we fall back to 1D linspace along $L$. The fingerprint is computed independently per view group: in the multi-camera setting, the seven cameras form three view groups (front, side, and rear) by shared grid shape, with cameras within a group stacked along $V$ and sharing one per-group fingerprint.

\subsubsection{Auxiliary channels} 
Because $\phi(\mathbf{x})$ is sparse and depends only on the block input, we attach two cheap auxiliary signals to close blind spots:
\begin{itemize}
    \item \textbf{Global channel:} The per-view-group sequence-mean of the block input, obtained by averaging along the token axis. It captures bulk latent drift that the sparse spatial sample may miss.
    \item \textbf{Condition channel:} The per-chunk action vector consumed by adaLN-Zero is flattened and appended as an additional fingerprint entry. Its shape is stable (fixed by the action dimensionality and the frames-per-chunk) and it changes once per generation, so the overhead is negligible. Including it lets the input-similarity metric react directly to per-step control maneuvers that would otherwise only propagate through the block-0 cascade. Other conditioning signals (dynamic-object and lane embeddings injected via cross-attention, text via cross-attention) have variable padding across chunks and slower change timescales; we leave them to the cascade and anchor-block mechanisms described in Section~\ref{sec:method:safety}.
\end{itemize}
All fingerprint entries are flattened before metric computation.

\subsubsection{Metric 1: Cosine similarity (global direction)}
\begin{equation}
\label{eq:cosine}
s_{\cos} = \frac{\phi(\mathbf{x}^{(n)}_{t,b-1}) \cdot \phi(\mathbf{x}^{(n-1)}_{t,b-1})}{\|\phi(\mathbf{x}^{(n)}_{t,b-1})\| \cdot \|\phi(\mathbf{x}^{(n-1)}_{t,b-1})\|},
\end{equation}
computed per fingerprint entry (each spatial view group, plus the global and action-condition channels) and aggregated by taking the minimum across all entries: $s_{\cos} = \min_{k} s^{(k)}_{\cos}$.

\subsubsection{Metric 2: Maximum token deviation (local outlier)}
\begin{equation}
\label{eq:deviation}
d_{\max} = \frac{\max\!\left|\phi(\mathbf{x}^{(n)}_{t,b-1}) - \phi(\mathbf{x}^{(n-1)}_{t,b-1})\right|}{\text{mean}\!\left|\phi(\mathbf{x}^{(n-1)}_{t,b-1})\right| + \epsilon},
\end{equation}
computed only over the per-group spatial fingerprints (the global and action-condition channels use cosine alone) and aggregated by taking the maximum across view groups: $d_{\max} = \max_{g} d^{(g)}_{\max}$.

\subsubsection{Skip decision} 
Block $b$ at denoising step $t$ is skipped if and only if both metrics pass:
\begin{equation}
\label{eq:skip}
\text{skip}(t, b) = \left(s_{\cos} \geq \tau_{\cos}(t, b)\right) \;\wedge\; \left(d_{\max} < \tau_{\text{dev}}\right),
\end{equation}
where $\tau_{\cos}(t, b)$ is an adaptive threshold (Section~\ref{sec:method:threshold}) and $\tau_{\text{dev}}$ is a fixed deviation threshold. The conservative aggregation (min-cosine over all fingerprint entries, max-deviation over spatial view groups) ensures that anomalous change in any single view group, in the global summary, or in the action-condition channel triggers recomputation.

\subsection{Adaptive Threshold}
\label{sec:method:threshold}

Rather than a fixed cosine threshold, X-Cache learns a per-position threshold from the block's own history. For each $(t, b)$ position, we maintain an exponential moving average of the observed cosine similarity:
\begin{equation}
\label{eq:ema}
\bar{s}_{t,b} \leftarrow \alpha \cdot s_{\cos}^{(n)} + (1 - \alpha) \cdot \bar{s}_{t,b},
\end{equation}
updated each time the test at $(t, b)$ is evaluated ($\alpha = 0.3$). The adaptive threshold is:
\begin{equation}
\label{eq:adaptive_th}
\tau_{\cos}(t,b) = \max\!\left(\tau_{\text{floor}},\;\bar{s}_{t,b} - m\right),
\end{equation}
where $m$ is a configurable margin and $\tau_{\text{floor}}$ is a hard quality floor. In our implementation, $\tau_{\text{floor}} = 0.95$ and $m = 0.02$.

Blocks with consistently high cross-chunk similarity accumulate high EMAs and their thresholds settle just below their typical similarity, maximizing skips. Blocks with volatile similarity remain conservative. The quality floor provides an absolute safety guarantee independent of history.

\subsection{Safety Mechanisms}
\label{sec:method:safety}

\paragraph{Condition visibility gap} External conditioning signals are injected inside each DiT block, not into the block input $\mathbf{x}$. The spatial fingerprint $\phi(\mathbf{x})$ therefore does not directly reflect condition changes. The action-condition channel from Section~\ref{sec:method:gating} lifts the action vector into fingerprint space, closing the gap for per-step controls. Dynamic-object and lane embeddings have variable padding across chunks and change on slower timescales, so including them directly in the fingerprint is neither shape-stable nor cost-effective; together with the text condition they are handled instead by the following mechanisms:

\begin{enumerate}
    \item \textbf{Denoising step 0 protection:} At $t=0$, the input latent is dominated by high-level noise, and conditioning signals have maximal relative influence on the block output. Noise is also resampled at each KV update cycle and new context from the previous chunk's output is embedded in the noisy input, so consecutive generations naturally exhibit low cosine similarity at $t=0$. By default, X-Cache forces full computation at $t=0$. An optional relaxation mode applies a strict threshold $\tau^{\text{strict}}_0 = 0.999$, which is near-unity: even minor latent changes drop the similarity below 0.999, keeping step 0 effectively protected. Once step 0 computes with updated conditions, its output cascades through subsequent steps: each block's input is derived from the prior step's output, so fingerprints diverge from the cached versions and trigger recomputation where needed.
    \item \textbf{Anchor blocks ($F_n$):} The first $F_n$ blocks are unconditionally computed at all denoising steps (default $F_n = 1$). With $F_n = 1$, block 0 always processes the current conditioning via adaLN-Zero, and its changed output cascades through subsequent blocks' fingerprints. This provides a per-step guarantee independent of step 0 protection. The last $B_n$ blocks can similarly be designated as tail anchors (default $B_n = 0$).
    \item \textbf{KV update frame protection:} The generation step whose clean latent will be used for the KV update pass is critical: its KV projections are attended to by all future chunks. During this generation, X-Cache enters force-compute mode: all blocks compute fully, but the cache remains attached so that fingerprints and residuals are refreshed for the next generation. This ensures both (a) precise KV entries and (b) that the subsequent generation starts with fresh cache data, avoiding a double-heavy-frame penalty.
    \item \textbf{Maximum staleness:} A counter tracks consecutive skips per $(t, b)$ position. If it exceeds threshold $M$, the block is forced to recompute.
\end{enumerate}
\section{Experiments}
\label{sec:experiments}

\subsection{Settings}

\subsubsection{Hardware}
\label{sec:exp:hardware}
\begin{CJK*}{UTF8}{gbsn}
All of our experiments run on the Zhenwu (真武) 810E, an AI accelerator developed by Alibaba T-Head. In the remainder of this paper we will refer to the device as a Parallel Processing Unit (PPU). Each PPU integrates 96 GB of HBM2e on-chip memory and natively supports FP16, BF16, and INT8 with hardware acceleration. In the following experiments, all DiT forward passes are executed in BF16 on a single PPU.
\end{CJK*}

\subsubsection{Experiment Model}
\label{sec:exp:model}

We evaluate on X-World~\cite{x-world}, a controllable ego-centric multi-camera world model for autonomous driving built on WAN 2.2~\cite{wan2025}. X-World follows the latent video diffusion paradigm, combining a causal VAE with a DiT denoiser to generate synchronized 360\textdegree\ video from seven cameras at 12 FPS. During streaming autoregressive rollout, it generates video chunk by chunk: each chunk starts from Gaussian noise, run 4-step denoising, and updates a fixed-size rolling KV cache with FIFO eviction to support long-horizon simulation under bounded memory.

\subsubsection{Dataset}
\label{sec:exp:data}

We evaluate on an internal held-out test split drawn from the X-World training distribution, exercised under the interactive closed-loop inference protocol X-World is designed for. At each generation chunk the causal DiT attends only to one frame of initial 7-camera history, the conditioning that applies to that chunk alone (ego action state, dynamic-agent poses, static road-element annotations, a scene-level text caption), and the rolling KV cache of past chunks; the causal architecture forbids any look-ahead to future conditioning. The per-chunk actions are not known to the model in advance: they are replayed one chunk at a time from a recorded trajectory so each run is reproducible, which is indistinguishable from a live policy stream from the DiT's perspective. Each clip produces 264 frames (around 22 seconds at 12 FPS) of future multi-camera video, with no per-frame visual ground truth supplied during rollout. The split covers three scenario groups: 7 urban-street clips with dense surrounding traffic, pedestrians, and storefronts; 3 highway clips covering elevated urban express ring-road and ordinary motorway; and 3 urban u-turn clips where the ego vehicle executes sharp heading changes and consecutive chunks change the most across views. We use the u-turn group to stress-test the gating assumption under the largest cross-chunk motion in the split.

\subsubsection{Metrics}
\label{sec:exp:metrics}

Since X-Cache only modifies the DiT denoiser, we evaluate only metrics that directly reflect its effect on denoising. We do not include VAE encoding or decoding, data loading, post-processing, or inter-device transfer in any reported numbers. On compute side, we report the block skip rate, defined as the fraction of DiT block evaluations that reuse a cached residual during a rollout, excluding warmup chunks. We also show how this skip rate varies across denoising steps and block indices. For efficiency, we report the average per-chunk DiT wall-clock time on a single PPU, as well as the resulting DiT speedup relative to a full-compute run with the same seed, conditioning, and KV state. For fidelity, we compare each X-Cache rollout with its corresponding full-compute reference rollout. Because both branches share the same decoder, any difference comes only from the DiT side. We report three frame-level metrics on the decoded images, averaged over all seven cameras: PSNR for pixel-level fidelity, SSIM for structural fidelity, and LPIPS~\cite{lpips} for perceptual fidelity on deep features.

\subsubsection{Default Parameters}
\label{sec:exp:defaults}

Unless otherwise specified, X-Cache is run with the default hyperparameters given in Table~\ref{tab:xcache_defaults}.

\begin{table}[h]
\centering
\begin{tabular}{l l c}
\toprule
Symbol & Meaning & Default \\
\midrule
$F_n$              & Number of front anchor blocks (always computed)   & 1 \\
$B_n$              & Number of back anchor blocks (always computed)    & 0 \\
$W$                & Warmup chunks (all blocks fully computed)         & 1 \\
$\tau_{\text{floor}}$ & Adaptive-threshold quality floor                 & 0.97 \\
$m$                & Adaptive-threshold margin below EMA               & 0.02 \\
$\alpha$           & EMA update coefficient                            & 0.30 \\
$\tau_{\text{dev}}$ & Max-token deviation threshold                    & 2.00 \\
$M$                & Max consecutive skips per $(t,b)$                 & Off \\
KV-update-chunk protection & Force full compute on KV-update chunk     & On \\
Denoising step-0 protection & Force full compute at $t{=}0$            & Off \\
\bottomrule
\end{tabular}
\vspace{5pt}
\caption{Default X-Cache hyperparameters used throughout the experiments.}
\label{tab:xcache_defaults}
\end{table}

\subsection{Results}
\label{sec:exp:main}

\begin{figure}[h]
\centering
\includegraphics[width=\linewidth]{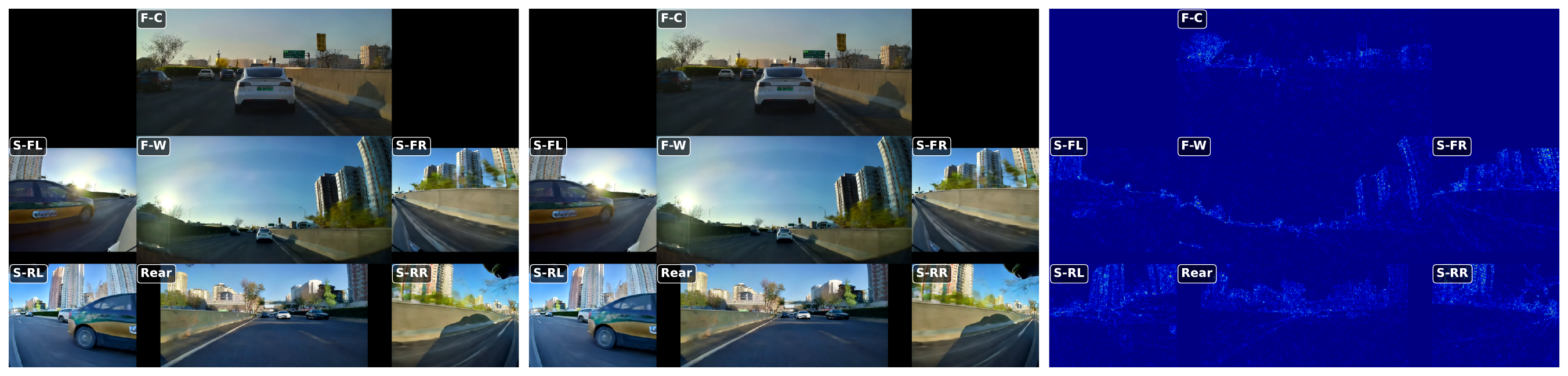}\\[2pt]
\includegraphics[width=\linewidth]{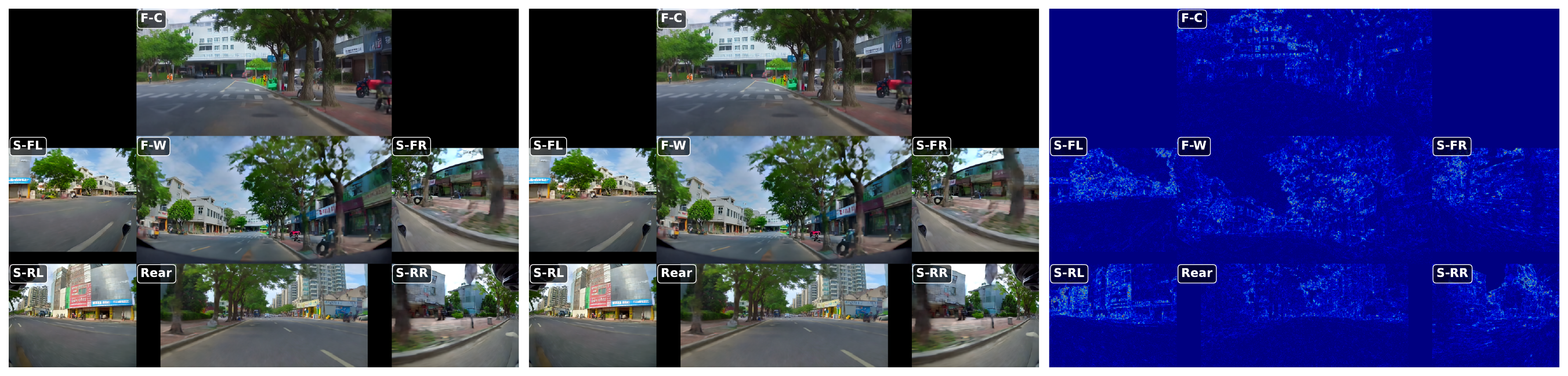}\\[2pt]
\includegraphics[width=\linewidth]{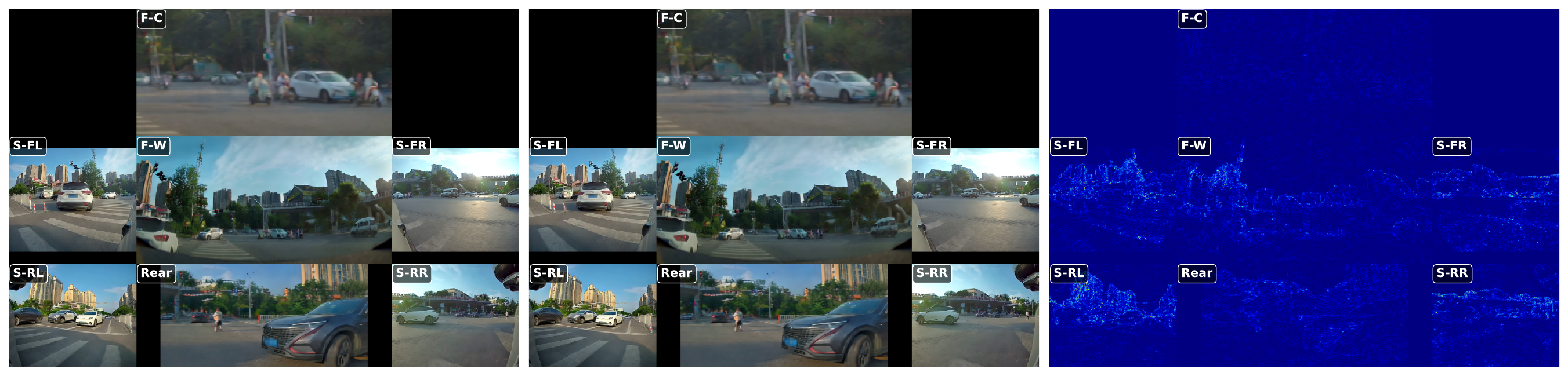}
\caption{Qualitative frame-level comparison in three scenarios. Rows, top to bottom: highway, urban street, and u-turn. Each row: \textit{left} baseline full-compute, \textit{middle} X-Cache under identical seed and conditioning, \textit{right} absolute pixel difference amplified $20\times$. Per-view labels match Table~\ref{tab:main}.}
\label{fig:qualitative}
\end{figure}

\begin{table}[h]
\centering
\small
\setlength{\tabcolsep}{4pt}
\renewcommand{\arraystretch}{1.05}
\begin{tabular}{@{}l l c c c c c c@{}}
\toprule
Scenario & Camera & PSNR (dB) $\uparrow$ & SSIM $\uparrow$ & LPIPS $\downarrow$ & Skip (\%) & DiT (s) & Speedup \\
\midrule
\multirow{8}{*}{Urban Street ($n{=}7$)}
 & F-C   & 53.83  & 0.9988 & 3.6e-4 & \multirow{8}{*}{71.4} & \multirow{8}{*}{1.392} & \multirow{8}{*}{2.65$\times$} \\
 & F-W   & 50.27  & 0.9987 & 4.3e-4 & & & \\
 & S-FL  & 49.49  & 0.9985 & 5.1e-4 & & & \\
 & S-FR  & 48.69  & 0.9984 & 5.2e-4 & & & \\
 & S-RL  & 48.59  & 0.9985 & 4.8e-4 & & & \\
 & S-RR  & 48.07  & 0.9985 & 5.2e-4 & & & \\
 & Rear  & 51.77  & 0.9986 & 4.7e-4 & & & \\
\rowcolor{gray!15}
 & 7-cam & 51.37  & 0.9990 & 3.3e-4 & & & \\
\midrule
\multirow{8}{*}{Highway ($n{=}3$)}
 & F-C   & 54.87  & 0.9989 & 2.6e-4 & \multirow{8}{*}{71.6} & \multirow{8}{*}{1.365} & \multirow{8}{*}{2.66$\times$} \\
 & F-W   & 54.38  & 0.9988 & 2.3e-4 & & & \\
 & S-FL  & 53.08  & 0.9987 & 2.8e-4 & & & \\
 & S-FR  & 52.20  & 0.9987 & 2.9e-4 & & & \\
 & S-RL  & 52.48  & 0.9987 & 2.5e-4 & & & \\
 & S-RR  & 51.90  & 0.9986 & 3.0e-4 & & & \\
 & Rear  & 53.42  & 0.9987 & 3.2e-4 & & & \\
\rowcolor{gray!15}
 & 7-cam & 54.67  & 0.9991 & 1.9e-4 & & & \\
\midrule
\multirow{8}{*}{U-turn ($n{=}3$)}
 & F-C   & 54.60  & 0.9987 & 4.3e-4 & \multirow{8}{*}{71.3} & \multirow{8}{*}{1.364} & \multirow{8}{*}{2.70$\times$} \\
 & F-W   & 51.79  & 0.9987 & 3.6e-4 & & & \\
 & S-FL  & 49.29  & 0.9985 & 4.6e-4 & & & \\
 & S-FR  & 49.18  & 0.9985 & 4.7e-4 & & & \\
 & S-RL  & 48.87  & 0.9985 & 4.0e-4 & & & \\
 & S-RR  & 48.82  & 0.9984 & 4.9e-4 & & & \\
 & Rear  & 52.51  & 0.9986 & 4.2e-4 & & & \\
\rowcolor{gray!15}
 & 7-cam & 52.04  & 0.9990 & 3.1e-4 & & & \\
\bottomrule
\end{tabular}
\vspace{3pt}
\caption{X-Cache vs.\ baseline full-compute reference. Per-camera rows are F-C (front centre), F-W (front wide), S-FL/FR/RL/RR (four sides), and Rear. The shaded "7-cam" row is computed on the stitched seven-view image used in the qualitative figures (not the arithmetic mean of the per-camera rows). Skip rate excludes warmup; DiT wall-clock is per-chunk on a single PPU, excluding the first chunk. Baseline DiT is 3.682\,s / 3.633\,s / 3.688\,s per chunk on urban / Highway / u-turn.}
\label{tab:main}
\end{table}

\begin{figure}[h]
\centering
\includegraphics[width=0.98\linewidth]{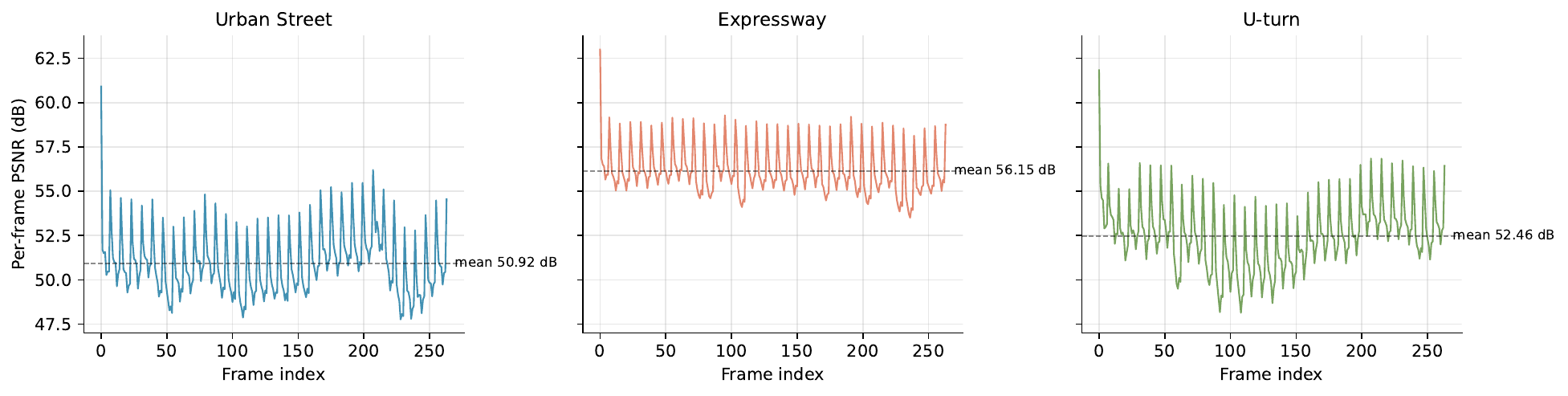}
\caption{Per-frame combined-7-view PSNR of X-Cache rollouts against their baseline full-compute references, on one representative session per scenario. Dashed line marks the per-session mean.}
\label{fig:psnr_per_frame}
\end{figure}

\begin{figure}[h]
\centering
\includegraphics[width=0.98\linewidth]{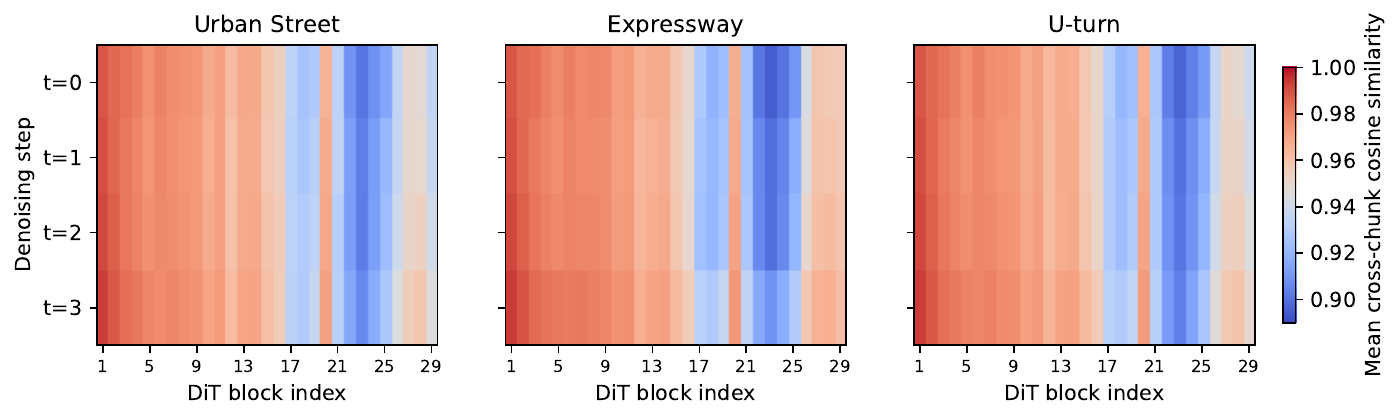}
\caption{Per-$(\text{step}, \text{block})$ cross-chunk cosine similarity, one panel per scenario, averaged over post-warmup chunks of a representative clip. Block~0 is omitted because the front anchor ($F_n{=}1$) never participates in the gate. Color scale is shared across the three panels.}
\label{fig:cosine_heatmap}
\end{figure}

\begin{figure}[h]
\centering
\includegraphics[width=0.85\linewidth]{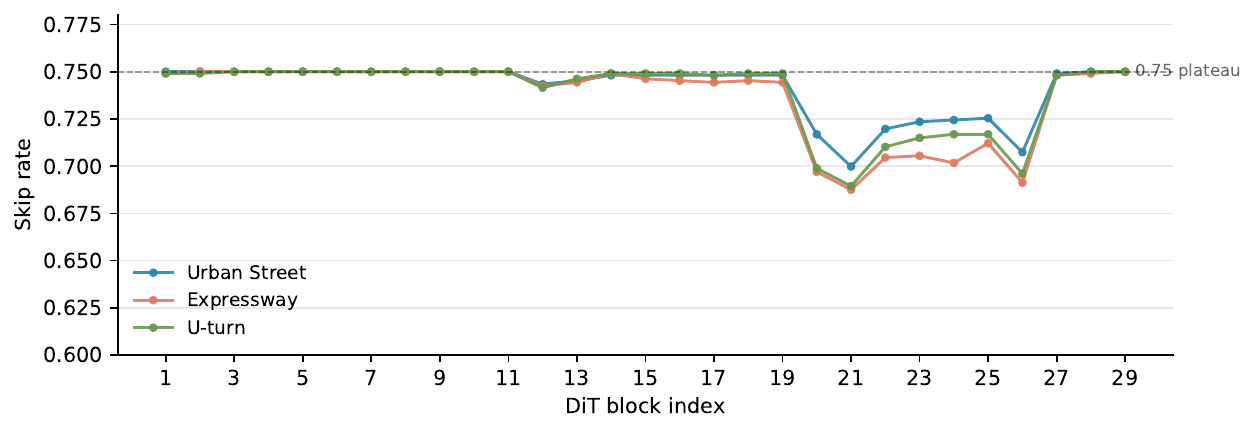}
\caption{Per-block skip rate, averaged across the four denoising steps and across post-warmup chunks of one representative clip per scenario. Block 0 (front anchor, $F_n{=}1$) is omitted. Dashed line marks the theoretical 0.75 plateau implied by $W{=}1$ warmup over the 4-chunk reuse window.}
\label{fig:skip_per_block}
\end{figure}

X-Cache gives 2.65 to 2.70$\times$ DiT speedup across the three splits while keeping decoded quality in the imperceptible range (SSIM above 0.9990, LPIPS below 4e-4, 7-camera PSNR between 51 and 55 dB in Table~\ref{tab:main}). Scene complexity produces a small but real spread in quality, with the shaded 7-camera PSNR increasing from 51.37 dB for urban street to 52.04 dB for u-turn and 54.67 dB for highway. Skip rate and DiT time do not follow this spread: they stay within 0.3 percentage points and 30 ms, respectively, across all three splits. The quality ordering reflects how each scenario's decoded pixels absorb a common cache-induced latent perturbation, not a scenario-specific slowdown or tightening of the gate itself. The qualitative residuals in Figure~\ref{fig:qualitative} also agree with this: the three difference panels show the same sparsity pattern, with the residual sitting on lane edges and far-field foliage and disappearing at display scale even under 20 $\times$ amplification.

It is counter-intuitive that the u-turn split slightly outperforms the urban split, given that u-turn exhibits the largest cross-chunk motion in the dataset; this can be explained by two clip-level properties. First, the turn occupies only a minority of each u-turn clip, with most of the 22 seconds spent in straight driving or lane change, so the clip-averaged PSNR averages urban-like frames with a smaller stretch of genuine turn. Second, during the yaw the cameras register real-world motion blur across the lateral field. The blur smooths fine texture in the reference frames and reduces the high-frequency content that cache-induced latent perturbations would otherwise surface as pixel error, so the in-turn frames themselves are easier to match than a sharp urban frame of comparable complexity. The combined effect places the u-turn between urban and highway rather than below urban. Figure~\ref{fig:psnr_per_frame} shows the structure behind this average: the u-turn curve sits near the urban level outside the turn window and has a visible drop between frames 50 and 140 where the per-chunk fingerprint change exceeds the adaptive threshold more often, then recovers to the pre-turn level once the fingerprint settles.

The gate behaves nearly identically across scenarios: all three skip the same set of blocks. Figure~\ref{fig:cosine_heatmap} shows nearly identical heatmap shapes across panels. Blocks 1 to 19 remain above 0.95, whereas blocks 20 to 26 drop to around 0.90, and variation along the denoising-step axis stays within about 0.02 for any fixed block. Figure~\ref{fig:skip_per_block} shows the same partition at the decision level: blocks 1 to 19 stay on the 0.75 plateau, whereas blocks 20 to 26 fall to about 0.69, with all three scenarios following essentially the same curve. The 0.3 percentage-point spread in skip rate in Table~\ref{tab:main} therefore comes from small movements within this fixed set of blocks, rather than from scenario-dependent changes in which blocks are cacheable. The per-chunk reuse budget also binds before the cosine test does: even u-turn, which has the largest inter-chunk motion, does not fall to a lower skip rate.

Two further observations reinforce this interpretation. First, the per-frame PSNR traces show no drift at chunk boundaries in any of the three clips, consistent with KV-update protection remaining effective over the 22-second rollouts we test. Second, the per-camera PSNR ranking within each scenario (front-center highest, rear intermediate, and side cameras 2 to 6 dB lower) should not be interpreted as a cache-side signal, because the gate makes a single decision per $(t, b)$ cell using a 7-camera-aware fingerprint (Section ~\ref{sec:method:gating}), rather than separate decisions for each camera. When a single PSNR per scenario is needed, the shaded 7-camera column is therefore the appropriate value to report.
\section{Ablation}
\label{sec:ablation}

We keep all settings from Section~\ref{sec:exp:defaults} fixed and vary one protection mechanism or hyperparameter at a time on a single held-out rollout (another highway clip with 264 frames). Table~\ref{tab:ablation} reports the same fidelity metrics as Table~\ref{tab:main}, together with per-chunk DiT wall-clock time and DiT speedup relative to the no-cache baseline on the same clip. Decoder and I/O are excluded, so the speedup column reflects only the portion affected by X-Cache.

\begin{table}[h]
\centering
\small
\begin{tabular}{l c c c c c c}
\toprule
Configuration & PSNR (dB) $\uparrow$ & SSIM $\uparrow$ & LPIPS $\downarrow$ & Skip (\%) & DiT (s) & Speedup \\
\midrule
Baseline (no cache)                       & --     & --     & --       & --   & 3.637 & 1.00$\times$ \\
\midrule
\textbf{Default} (step-0 off, $\tau_{\text{floor}}{=}0.97$) & 53.384 & 0.9990 & 2.0e-4   & 71.3 & 1.406 & 2.59$\times$ \\
$\;+$ Step-0 protection                    & 53.389 & 0.9990 & 2.0e-4   & 53.5 & 1.975 & 1.84$\times$ \\
$\;-$ KV-update protection                 & 21.461 & 0.8067 & 1.77e-1  & 62.8 & 1.670 & 2.18$\times$ \\
$\;-$ Front anchor ($F_n{=}0$)$^\dagger$   & 53.622 & 0.9991 & 2.0e-4   & 55.4 & 1.902 & 1.91$\times$ \\
$\tau_{\text{floor}} \in \{0.90,\dots,0.96\}^\dagger$ & 53.37--53.40 & 0.9990 & 2.0e-4 & 53.5 & 1.96--1.98 & 1.84$\times$ \\
\bottomrule
\end{tabular}
\vspace{3pt}
\caption{X-Cache ablations on the single 264-frame clip. $^\dagger$ The $F_n$ row and the $\tau_{\text{floor}}$ sweep keep step-0 protection \emph{on}, so the effect of each knob is isolated rather than entangled with the step-0 decision; the \textbf{Default} and $\;+$Step-0 rows already measure that decision on its own.}
\label{tab:ablation}
\end{table}

\subsection{Denoising step-0 protection} 
By default, step-0 skipping is enabled. Enabling step-0 protection removes all step-0 skips, reducing the overall skip rate from 71.3\% to 53.5\% and increasing per-chunk DiT wall-clock time from 1.41 s to 1.98 s (speedup from 2.59$\times$ to 1.84$\times$). On this clip, however, quality remains unchanged to the fourth decimal place of SSIM and LPIPS. In practice, whenever reuse is valid, the step-0 residual fingerprints already lie above the adaptive threshold, so the cosine-similarity gate alone is sufficient to suppress the cases of step-0 reuse that would otherwise matter. We nevertheless retain this option because the step-0 residual initializes all downstream blocks. Under larger distribution shift (for example, night, heavy rain, or sudden trajectory changes) or in longer rollouts where small errors may accumulate, forcing step-0 recomputation provides an additional safety margin without requiring any other change.

\subsection{KV-update-chunk protection}
This is the only setting whose removal causes a substantial loss in quality. Allowing the gate to skip tokens on the chunk where the rolling KV cache is updated reduces PSNR from 53.4 dB to 21.5 dB and increases LPIPS by three orders of magnitude, while improving skip rate by only 9 percentage points and changing DiT speedup from 2.59$\times$ to 2.18$\times$. This trade-off is clearly unfavorable, so KV-update-chunk protection remains enabled.

\subsection{Front anchor $F_n=1$}
Setting $F_n=0$ increases skip rate by 1.9 percentage points and saves about 70 ms of DiT time per chunk at essentially the same PSNR. However, block~0 defines the residual basis on which all later blocks build. If block~0 is skipped during a KV-update cycle, downstream error propagation is no longer naturally bounded. Our current clips do not strongly expose this failure mode, but we set the default to $F_n=1$ so that deployment on rarer scenarios does not rely on these clips being benign.

\subsection{Adaptive-threshold floor}
Sweeping $\tau_{\text{floor}}$ from 0.90 to 0.96 leaves all reported metrics unchanged at the precision shown here. Under the current workload, the floor is effectively inactive for two reasons. First, the measured per-chunk cosine-similarity distribution is heavily concentrated at 1.0 (both the median and p75 are 1.0 across all 7 urban-street clips), so very few tokens fall in the 0.94--0.97 range where lowering the floor would alter the skip decision. Second, the effective threshold is $\max(\tau_{\text{floor}}, \mathrm{EMA}_{\cos} - m)$, and on most $(t, b)$ cells the EMA-minus-margin term lies around 0.94--0.95. As a result, any $\tau_{\text{floor}}$ below about 0.94 is dominated by the EMA term rather than by the floor itself. The sweep therefore shows that the gate is insensitive to this parameter on the current test distribution, but not that it will remain insensitive on data farther from training. The floor still defines the minimum threshold the gate may fall to when the cosine-similarity distribution shifts, so we keep it configurable and choose a conservative default.

\section{Limitations}
\label{sec:limitations}

All rollouts we report are 22-second clips from an internal held-out split of the X-World training distribution, across urban street, highway, and u-turn. We have not measured X-Cache on longer horizons or on clips outside that distribution such as night-time, adverse weather, aggressive driving, or sustained highway cruising; in those scenario cross-chunk similarity patterns may shift and the adaptive EMA threshold would need time to recalibrate before the reported skip rate and quality numbers would apply. The defaults in Table~\ref{tab:xcache_defaults} are also tuned on a single held-out clip and are chosen to be safe rather than aggressive: the 7-camera PSNR settles around 53 dB with SSIM above 0.999 and LPIPS below $4\times 10^{-4}$, well inside the imperceptible range, so a direct parameter sweep (a lower $\tau_{\text{floor}}$, a relaxed $\tau_{\text{dev}}$, or dropping the front anchor on blocks whose cross-chunk cos stays high) should trade a small amount of the current quality margin for more skip rate and more wall-clock. We have not mapped out this Pareto frontier on our test set.

\section{Conclusion}
\label{sec:conclusion}

X-Cache targets a redundancy axis that cross-step caching methods do not reach: block-level residual reuse across consecutive generation chunks of a few-step autoregressive video diffusion model. Validated on X-World, a production driving world model, the default configuration reuses 71\% of DiT block evaluations and delivers 2.6--2.7$\times$ DiT wall-clock speedup against the no-cache baseline with no visible quality degradation. The behaviour holds across urban streets, highway driving, and u-turn rollouts, and the underlying cross-chunk cosine pattern is driven by block position in the DiT rather than by scene content. Our ablations identify KV-update-chunk protection as the one hard safety requirement and treat step-0 protection, the front anchor, and the adaptive-threshold floor as soft knobs that the current workload does not exercise but that the design retains as margins against distribution shift, long-horizon drift, and extreme operating conditions not covered by the current test split. Other few-step interactive AR world models that share this template, such as WorldPlay~\cite{WorldPlay}, are natural next extension targets, although the achievable speedup and quality floor under their larger and more abrupt action transitions still need to be verified.

\newpage
\section{Contributors}
\label{sec:contributors}
We extend our sincere gratitude to the entire team for their dedication and hard work. This project is a testament to our collective effort in pushing the boundaries of world model research and engineering.

\vspace{1em}

\begin{description}
    \item[Advisors and Project Leads:] Yu Zhang, Boyang Wang, Linkun Xu, Siyuan Lu, Bo Tian, Xianming Liu
    \item[Contributors:] Yixiao Zeng$^*$, Jianlei Zheng, Chaoda Zheng, Shijia Chen, Mingdian Liu, Tongping Liu, Tengwei Luo
\end{description}

\noindent $^*$Core contributor
\newpage
\bibliography{ref}
\bibliographystyle{plain}

\end{document}